# Staple: Complementary Learners for Real-Time Tracking

Luca Bertinetto Jack Valmadre Stuart Golodetz Ondrej Miksik Philip H.S. Torr
University of Oxford
{name.surname}@eng.ox.ac.uk

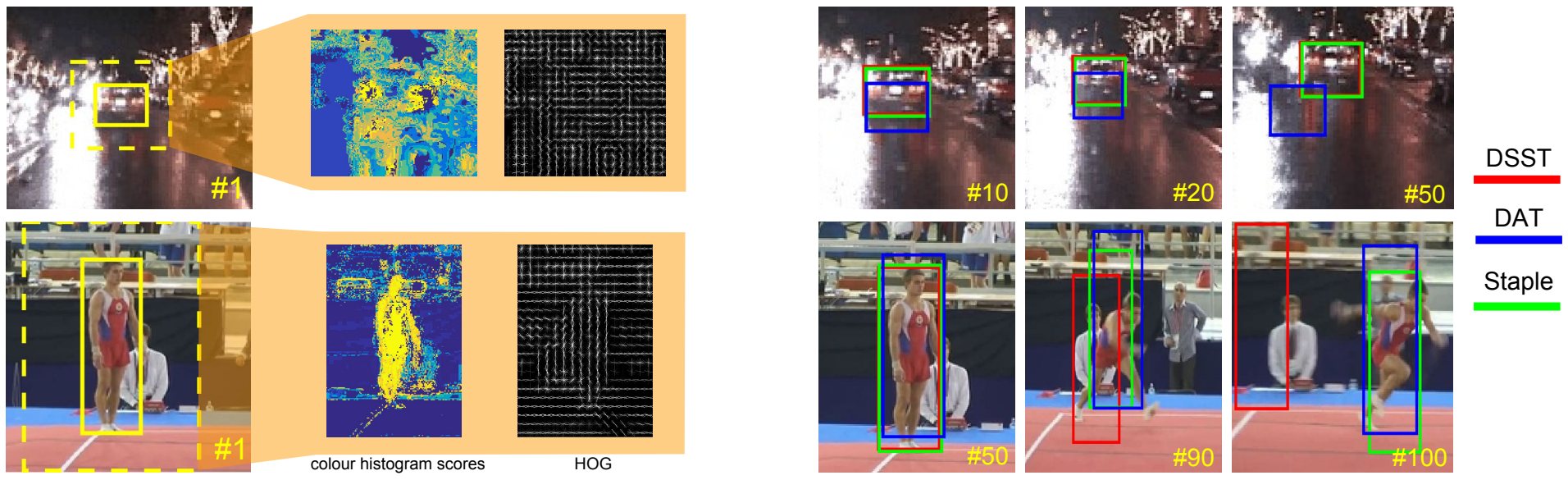

Figure 1: Sometimes colour distributions are not enough to discriminate the target from the background. Conversely, template models (like HOG) depend on the spatial configuration of the object and perform poorly when this changes rapidly. Our tracker *Staple* can rely on the strengths of both template and colour-based models. Like DSST [10], its performance is not affected by non-distinctive colours (top). Like DAT [33], it is robust to fast deformations (bottom).

## Abstract

*Correlation Filter-based trackers have recently achieved excellent performance, showing great robustness to challenging situations exhibiting motion blur and illumination changes. However, since the model that they learn depends strongly on the spatial layout of the tracked object, they are notoriously sensitive to deformation. Models based on colour statistics have complementary traits: they cope well with variation in shape, but suffer when illumination is not consistent throughout a sequence. Moreover, colour distributions alone can be insufficiently discriminative. In this paper, we show that a simple tracker combining complementary cues in a ridge regression framework can operate faster than 80 FPS and outperform not only all entries in the popular VOT14 competition, but also recent and far more sophisticated trackers according to multiple benchmarks.*

## 1. Introduction

We consider the widely-adopted scenario of short-term, single-object tracking, in which the target is only specified in the first frame (using a rectangle). *Short-term* implies that re-detection should not be necessary. The key challenge of tracking an unfamiliar object in video is to be robust to changes in its appearance. The task of tracking unfamiliar objects, for which training examples are not available in advance, is interesting because in many situations it is not feasible to obtain such a dataset. It is advantageous for the algorithm to perform above real-time for computationally intensive applications such as robotics, surveillance, video processing and augmented reality.

Since an object's appearance can vary significantly during a video, it is not generally effective to estimate its model from the first frame alone and use this single, fixed model to locate the object in all other frames. Most state-of-the-art algorithms therefore employ model adaptation to take advantage of information present in later frames. The simplest, most widespread approach is to treat the tracker's predictions in new frames as training data with which to update the model. The danger of learning from predictions is that small errors can accumulate and cause model drift. This is particularly likely to happen when the appearance of the object changes.

In this paper, we propose *Staple (Sum of Template And Pixel-wise LEarners)*, a tracker that combines two image patch representations that are sensitive to complementary factors to learn a model that is inherently robust to both colour changes and deformations. To maintain real-time speed, we solve two independent ridge-regression problems, exploiting the inherent structure of each representation. Compared to other algorithms that fuse the *predictions* of multiple models, our tracker combines the *scores* of two models in a dense translation search, enabling greater accuracy. A critical property of the two models is that their scores are similar in magnitude and indicative of their reliability, so that the prediction is dominated by the more confident.

We establish the surprising result that a simple combination of a Correlation Filter (using HOG features) and a global colour histogram outperforms many more complex trackers in multiple benchmarks while running at speeds in excess of 80 FPS.

## 2. Related Work

**Online learning and Correlation Filters.** Modern approaches to adaptive tracking often use an online version of an object detection algorithm. One approach that achieves strong results [39] and has an elegant formulation is Struck [16], which seeks to minimise the structured output objective for localisation [3]. However, the computation needed limits the number of features and training examples.

Correlation Filters instead minimise a least-squares loss for all circular shifts of the positive examples. Although this might seem a weaker approximation of the true problem, it enables the use of densely-sampled examples and high-dimensional feature images in real-time using the Fourier domain. Initially applied to adaptive tracking in grayscale images by Bolme *et al*. [5], their extension to multiple feature channels [4, 17, 22, 18] and therefore HOG features [7] enabled the technique to achieve state-of-the-art performance in VOT14 [24]. The winner of the challenge, DSST [8], incorporated a multi-scale template for Discriminative Scale-Space Tracking using a 1D Correlation Filter. One deficiency of Correlation Filters is that they are constrained to learn from *all* circular shifts. Several recent works [12, 23, 9] have sought to resolve this issue, and the Spatially Regularised (SRDCF) [9] formulation in particular has demonstrated excellent tracking results. However, this was achieved at the cost of real-time operation.

**Robustness to deformation.** Correlation Filters are inherently confined to the problem of learning a rigid template. This is a concern when the target experiences shape deformation in the course of a sequence. Perhaps the simplest method to achieve robustness to deformation is to adopt a representation that is insensitive to shape variation. Image histograms have this property, because they discard the position of every pixel. In fact, histograms can be considered orthogonal to Correlation Filters, since a Correlation Filter is *learnt from* circular shifts, whereas a histogram is *invariant to* circular shifts. However, histograms alone are often insufficient to discriminate the object from the background. While colour histograms were used in many early approaches to object tracking [32, 31], they have only recently been demonstrated to be competitive in modern benchmarks in the Distractor-Aware Tracker (DAT) [33], which uses adaptive thresholding and explicit suppression of regions with similar colours. In general, histograms may be constructed from any discrete-valued feature, including local binary patterns and quantised colours. For a histogram to provide robustness to deformation, the feature must be insensitive to the local changes that arise.

The chief alternative way to achieve robustness to deformation is to learn a deformable model. We believe it is ambitious to learn a deformable model from a single video in which the only supervision is the location in the first frame, and therefore adopt a simple bounding box. While our method outperforms recent sophisticated parts-based models [6, 40] in benchmarks, deformable models have a richer representation that these evaluations do not necessarily reward. Our single-template tracker could be considered a component with which to construct a parts-based model.

Rather than use a deformable model, HoughTrack [14] and PixelTrack [11] accumulate votes from each pixel and then use the pixels that voted for the winning location to estimate the object's extent. However, these methods are yet to demonstrate competitive benchmark performances.

**Schemes to reduce model drift.** Model drift is a result of learning from inaccurate predictions. Several works have aimed to prevent drift by modifying the training strategy rather than improving the predictions. TLD [21] and PROST [34] encode rules for additional supervision based on optical flow and a conservative appearance model. Other approaches avoid or delay making hard decisions. MILTrack [1] uses Multiple-Instance Learning to train with bags of positive examples. Supančič and Ramanan [35] introduce self-paced learning for tracking: they solve for the optimal trajectory keeping the appearance model, then update the model using the most confident frames, and repeat. Grabner *et al*. [15] treat tracking as online semi-supervised boosting, in which a classifier learnt in the first frame provides an anchor for the labels assigned to examples in later frames. Tang *et al*. [36] apply co-training to tracking, learning two independent SVMs that use different features and then obtaining hard negatives from the combined scores. Of these methods, only MILTrack and TLD are found in current benchmarks, and neither has strong results.

**Combining multiple estimates.** Another strategy widely adopted to mitigate inaccurate predictions is to combine the estimates of an *ensemble* of methods, so that the weaknesses of the trackers are reciprocally compensated. In [27, 28], Kwon *et al*. make use of complementary basic trackers, built by combining different observation models and motion models, and then integrate their estimates in a sampling framework. Similarly, [38] combines five independent trackers using a factorial HMM, modelling both the object trajectory and the reliability of each tracker across time. Rather than using trackers of different types, the Multi-Expert Entropy Minimisation (MEEM) tracker [41] maintains a collection of past models and chooses the prediction of one according to an entropy criterion. We differ from these approaches in that *a)* both of our models are learnt in a common framework (specifically, ridge regres-sion), and *b)* this enables us to directly combine the scores of the two models in a dense search.

**Long-term tracking with re-detection.** Several recent works have adopted Correlation Filters for the problem of long-term tracking, where the performance of an algorithm will be greatly improved by its ability to re-detect the object. The Long-term Correlation Tracker (LCT) [30] augments a standard Correlation Filter tracker with an additional Correlation Filter for confidence estimation and a random forest for re-detection, both of which are only updated in confident frames. The Multi-Store Tracker (MUSTer) [20] maintains a long-term memory of SIFT keypoints for the object and background, using keypoint matching and MLESAC to locate the object. The confidence of the long-term memory is estimated using the number of inliers, and occlusions can be determined by considering the number of background keypoints that are located inside the rectangle. Since we consider mostly short-term benchmarks, and these long-term trackers are meta-algorithms that are built upon a short-term tracker, there is little value in a comparison. Note that the TLD [21] and self-paced learning [35] algorithms also incorporate some aspects that are well-suited to the long-term tracking problem.

# 3. Proposed Approach

## 3.1. Formulation and motivation

We adopt the tracking-by-detection paradigm, in which, in frame $t$, the rectangle $p_t$ that gives the target location in image $x_t$ is chosen from a set $\mathcal{S}_t$ to maximise a score:

$$p_t = \arg\max_{p \in \mathcal{S}_t} f\left(T(x_t, p); \theta_{t-1}\right) \ . \tag{1}$$

The function $T$ is an image transformation such that $f(T(x, p); \theta)$ assigns a score to the rectangular window $p$ in image $x$ according to the model parameters $\theta$. The model parameters should be chosen to minimise a loss function $L(\theta; \mathcal{X}_t)$ that depends on the previous images and the location of the object in those images $\mathcal{X}_t = \{(x_i, p_i)\}_{i=1}^t$:

$$\theta_t = \arg\min_{\theta \in \mathcal{Q}} \{L(\theta; \mathcal{X}_t) + \lambda R(\theta)\} \ . \tag{2}$$

The space of model parameters is denoted $\mathcal{Q}$. We use a regularisation term $R(\theta)$ with relative weight $\lambda$ to limit model complexity and prevent over-fitting. The location $p_1$ of the object in the first frame is given. To achieve real-time performance, the functions $f$ and $L$ must be chosen not only to locate the object reliably and accurately, but also such that the problems in (1) and (2) can be solved efficiently.

We propose a score function that is a linear combination of template and histogram scores:

$$f(x) = \gamma_{\text{tmpl}} f_{\text{tmpl}}(x) + \gamma_{\text{hist}} f_{\text{hist}}(x) \ . \tag{3}$$

The template score is a linear function of a $K$-channel feature image $\phi_x : \mathcal{T} \to \mathbb{R}^K$, obtained from $x$ and defined on a finite grid $\mathcal{T} \subset \mathbb{Z}^2$:

$$f_{\text{tmpl}}(x; h) = \textstyle\sum_{u \in \mathcal{T}} h[u]^T \phi_x[u] \ . \tag{4}$$

In this, the weight vector (or *template*) $h$ is another $K$-channel image. The histogram score is computed from an $M$-channel feature image $\psi_x : \mathcal{H} \to \mathbb{R}^M$, obtained from $x$ and defined on a (different) finite grid $\mathcal{H} \subset \mathbb{Z}^2$:

$$f_{\text{hist}}(x; \beta) = g(\psi_x; \beta) \ . \tag{5}$$

Unlike the template score, the histogram score is invariant to spatial permutations of its feature image, such that $g(\psi) = g(\Pi\psi)$ for any permutation matrix $\Pi$. We adopt a linear function of the (vector-valued) average feature pixel

$$g(\psi; \beta) = \beta^T \left(\tfrac{1}{|\mathcal{H}|} \textstyle\sum_{u \in \mathcal{H}} \psi[u]\right) \ , \tag{6}$$

which can also be interpreted as the average of a scalar-valued score image $\zeta_{(\beta,\psi)}[u] = \beta^T \psi[u]$

$$g(\psi; \beta) = \tfrac{1}{|\mathcal{H}|} \textstyle\sum_{u \in \mathcal{H}} \zeta_{(\beta,\psi)}[u] \ . \tag{7}$$

To enable efficient evaluation of the score function in dense sliding-window search, it is important that both feature transforms commute with translation $\phi_{T(x)} = T(\phi_x)$. Not only does this mean that feature computation can be shared by overlapping windows, but also that the template score can be computed using fast routines for convolution, and that the histogram score can be obtained using a single integral image. Further acceleration is possible if the histogram weight vector $\beta$ or feature pixels $\psi[u]$ are sparse.

The parameters of the overall model are $\theta = (h, \beta)$, since the coefficients $\gamma_{\text{tmpl}}$ and $\gamma_{\text{hist}}$ can be considered implicit in $h$ and $\beta$. The training loss that will be optimised to choose parameters is assumed to be a weighted linear combination of per-image losses:

$$L(\theta, \mathcal{X}_T) = \textstyle\sum_{t=1}^T w_t \ell(x_t, p_t, \theta) \ . \tag{8}$$

Ideally, the per-image loss function should be of the form

$$\ell(x, p, \theta) = d(p, \arg\max_{q \in \mathcal{S}} f(T(x, q); \theta)) \ , \tag{9}$$

in which $d(p, q)$ defines the cost of choosing rectangle $q$ when the correct rectangle is $p$. Although this function is non-convex, structured output learning can be used to optimise a bound on the objective [3], and this is the basis of Struck [16]. However, the optimisation problem is computationally expensive, limiting the number of features and training examples that can be used. Correlation Filters, by contrast, adopt a simplistic least-squares loss, but are able to learn from a relatively large number of training examples using quite high-dimensional representations by considering circular shifts of the feature image as examples.

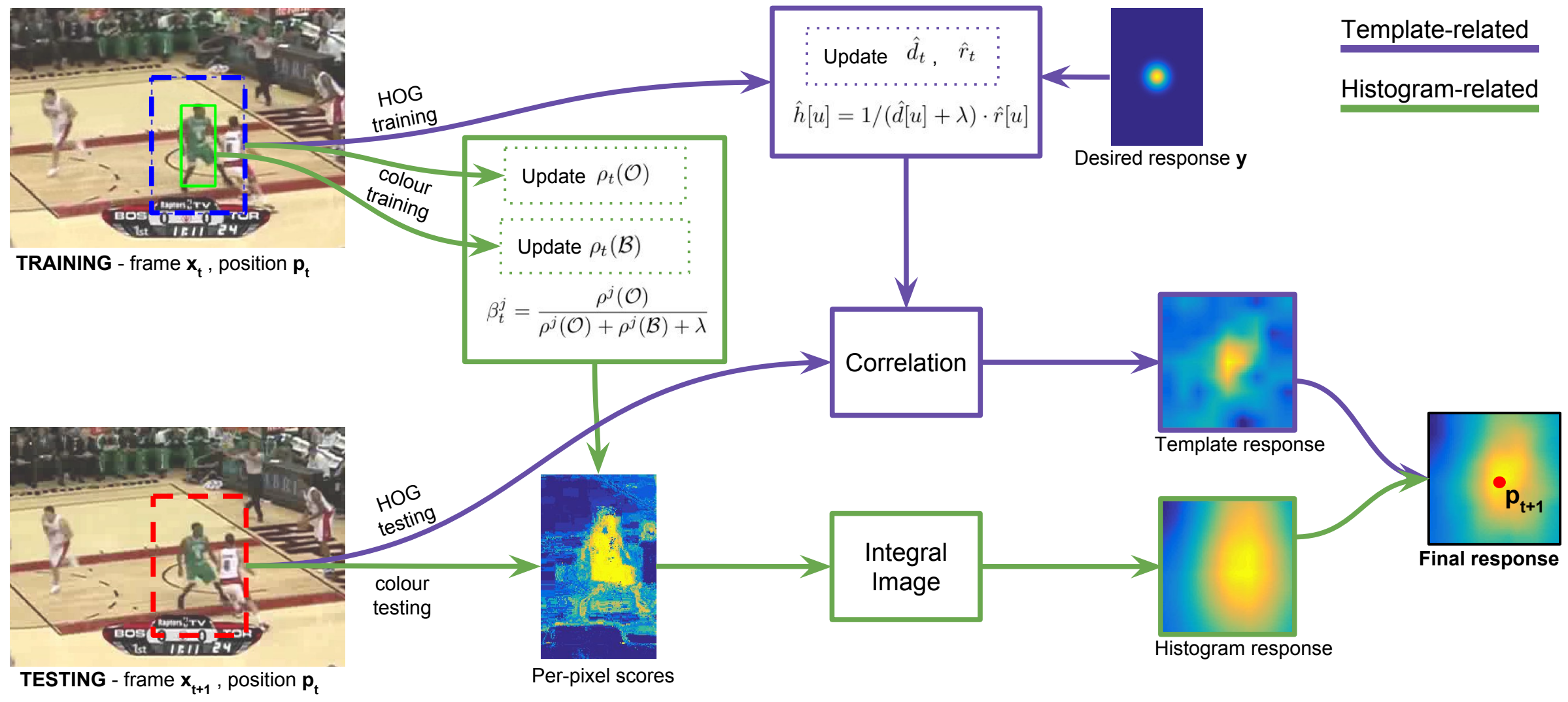

Figure 2: **Template-related.** In frame $t$, a training patch represented using HOG features is extracted at the estimated location $p_t$ and used to update the denominator $\hat{d}_t$ and the numerator $\hat{r}_t$ of the model $\hat{h}_t$ in (21). In frame $t+1$, features for the testing patch $\phi_{T(x_{t+1}, p_t)}$ are extracted around the location in the previous image $p_t$ and convolved with $\hat{h}_t$ in (4) to obtain the dense template response. **Histogram-related.** In frame $t$, foreground and background regions (relative to the estimated location) are used to update the frequencies of each colour bin $\rho_t(\mathcal{O})$ and $\rho_t(\mathcal{B})$ in (26). These frequencies enable us to compute the updated weights $\beta_t$. In frame $t+1$, a per-pixel score is computed in a search area centred at the position in the previous image, which is then used to compute the dense histogram response efficiently using an integral image (7). The final response is obtained with (3) and the new location $p_{t+1}$ of the target is estimated at its peak. Best viewed in colour.

(This requires the property that the feature transform commutes with translation.) This approach has achieved strong results in tracking benchmarks [18, 8] whilst maintaining high frame-rates.

It may at first seem counter-intuitive to consider $f_{\text{hist}}$ distinct from $f_{\text{tmpl}}$, when it is, in fact, a special case of $f_{\text{tmpl}}$ with $h[u] = \beta$ for all $u$. However, a uniform template such as this would not be learnt from circular shifts, since the score that is obtained using a uniform template is *invariant* to circular shifts. The histogram score may thus be understood to capture an aspect of the object appearance that is lost when considering circular shifts.

To retain the speed and efficacy of the Correlation Filter without ignoring the information that can be captured by a permutation-invariant histogram score, we propose to learn our model by solving two independent ridge regression problems:

$$\begin{aligned} h_t &= \arg\min_h \left\{ L_{\text{tmpl}}(h; \mathcal{X}_t) + \tfrac{1}{2}\lambda_{\text{tmpl}} \|h\|^2 \right\} \\ \beta_t &= \arg\min_\beta \left\{ L_{\text{hist}}(\beta; \mathcal{X}_t) + \tfrac{1}{2}\lambda_{\text{hist}} \|\beta\|^2 \right\} \end{aligned} \quad (10)$$

The parameters $h$ can be obtained quickly using the Correlation Filter formulation. While the dimension of $\beta$ may be less than that of $h$, it may still be more expensive to solve for, since it cannot be learnt with circular shifts and therefore requires the inversion of a general matrix rather than a circulant matrix. The fast optimisation of the parameters $\beta$ will be covered later in this section.

Finally, we take a convex combination of the two scores, setting $\gamma_{\text{tmpl}} = 1 - \alpha$ and $\gamma_{\text{hist}} = \alpha$, where $\alpha$ is a parameter chosen on a validation set. We hope that since the parameters of both score functions will be optimised to assign a score of 1 to the object and 0 to other windows, the magnitudes of the scores will be compatible, making a linear combination effective. Figure 2 is a visual representation of the overall learning and evaluation procedure.

### 3.2. Online least-squares optimisation

Two advantages of adopting a least-squares loss and a quadratic regulariser are that the solution can be obtained in closed form and the memory requirements do not grow with the number of examples. If $L(\theta; \mathcal{X})$ is a convex quadratic function of the score $f(x; \theta)$ and $f(x; \theta)$ is linear in the model parameters $\theta$ to preserve convexity, then there exists a matrix $A_t$ and a vector $b_t$ such that

$$L(\theta; \mathcal{X}_t) + \lambda \|\theta\|^2 = \tfrac{1}{2}\theta^T (A_t + \lambda I)\theta + b_t^T \theta + \text{const.} \quad (11)$$

and these are sufficient to determine the solution $\theta_t = (A_t + \lambda I)^{-1} b_t$, regardless of the size of $\mathcal{X}_t$. If we adopt a recursive definition of the loss function

$$L(\theta; \mathcal{X}_t) = (1 - \eta) L(\theta; \mathcal{X}_{t-1}) + \eta \ell(x_t, p_t, \theta) \quad (12)$$

with adaptivity rate $\eta$, then we can simply maintain

$$\begin{aligned} A_t &= (1-\eta)A_{t-1} + \eta A'_t \\ b_t &= (1-\eta)b_{t-1} + \eta b'_t \end{aligned} \tag{13}$$

in which $A'_t$ and $b'_t$ define the per-image loss via

$$\ell(x_t, p_t, \theta) = \tfrac{1}{2}\theta^T A'_t \theta + \theta^T b'_t + \text{const.} \tag{14}$$

Note that $A_t$ denotes the parameters estimated from frames 1 to $t$, whereas $A'_t$ denotes the parameters estimated from frame $t$ alone. We will be consistent in this notation.

These parameters, which are sufficient to obtain a solution, are generally economical to compute and store if the number of features (the dimension of $\theta$) is small or the matrix is redundant (e.g. sparse, low rank or Toeplitz). This technique for adaptive tracking with Correlation Filters using circulant matrices was pioneered by Bolme *et al.* [5].

### 3.3. Learning the template score

Under a least-squares Correlation Filter formulation, the per-image loss is

$$\ell_{\text{tmpl}}(x, p, h) = \left\| \textstyle\sum_{k=1}^{K} h^k \star \phi^k - y \right\|^2 \tag{15}$$

where $h^k$ refers to channel $k$ of multi-channel image $h$, $\phi$ is short for $\phi_{T(x,p)}$, $y$ is the desired response (typically a Gaussian function with maximum value 1 at the origin), and $\star$ denotes periodic cross-correlation. This corresponds to linear regression from the circular shift of $\phi$ by $\delta$ pixels to the value $y[\delta]$ with a quadratic loss. Using $\hat{x}$ to denote the Discrete Fourier Transform $Fx$, the minimiser of the regularised objective $\ell_{\text{tmpl}}(x, p, h) + \lambda\|h\|^2$ is obtained [22]

$$\hat{h}[u] = (\hat{s}[u] + \lambda I)^{-1}\hat{r}[u] \tag{16}$$

for all $u \in \mathcal{T}$, where $\hat{s}[u]$ is a $K \times K$ matrix with elements $\hat{s}^{ij}[u]$ and $\hat{r}[u]$ is a $K$-dimensional vector with elements $\hat{r}^i[u]$. Treating $s^{ij}$ and $r^i$ as signals, these are defined

$$s^{ij} = \phi^j \star \phi^i \;\; , \qquad\qquad r^i = y \star \phi^i \tag{17}$$

or, in the Fourier domain, using $^*$ to denote conjugation and $\odot$ for element-wise multiplication,

$$\hat{s}^{ij} = (\hat{\phi}^j)^* \odot \hat{\phi}^i \;\; , \qquad \hat{r}^{ij} = (\hat{y})^* \odot \hat{\phi}^i \;\; . \tag{18}$$

In practice, Hann windowing is applied to the signals to minimise boundary effects during learning. Instead of computing (16), we adopt the approximation found in the DSST code [8]

$$\hat{h}[u] = 1/(\hat{d}[u] + \lambda) \cdot \hat{r}[u] \;\; . \tag{19}$$

where $\hat{d}[u] = \text{tr}(\hat{s}[u])$ or

$$\hat{d} = \textstyle\sum_{i=1}^{K} (\hat{\phi}^i)^* \odot \hat{\phi}^i \tag{20}$$

This enables the algorithm to remain fast with a significant number of feature channels, since it is not necessary to factorise a matrix per pixel. The online version update is

$$\begin{aligned} \hat{d}_t &= (1-\eta_{\text{tmpl}})\hat{d}_{t-1} + \eta_{\text{tmpl}}\hat{d}'_t \\ \hat{r}_t &= (1-\eta_{\text{tmpl}})\hat{r}_{t-1} + \eta_{\text{tmpl}}\hat{r}'_t \end{aligned} \tag{21}$$

where $\hat{d}'$ and $\hat{r}'$ are obtained according to (20) and (18) respectively. For a template of $m = |\mathcal{T}|$ pixels and $K$ channels, this can be performed in $O(Km \log m)$ time and the sufficient statistics $\hat{d}$ and $\hat{r}$ require $O(Km)$ memory.

### 3.4. Learning the histogram score

Ideally, the histogram score should be learnt from a set of examples taken from each image, including the correct position as a positive example. Let $\mathcal{W}$ denote a set of pairs $(q, y)$ of rectangular windows $q$ and their corresponding regression target $y \in \mathbb{R}$, including the positive example $(p, 1)$. The per-image loss is then $\ell_{\text{hist}}(x, p, \beta) =$

$$\textstyle\sum_{(q,y)\in\mathcal{W}} \left(\beta^T \left[\sum_{u\in\mathcal{H}} \psi_{T(x,q)}[u]\right] - y\right)^2 . \tag{22}$$

For an $M$-channel feature transform $\psi$, the solution is obtained by solving an $M \times M$ system of equations, which requires $O(M^2)$ memory and $O(M^3)$ time. If the number of features is large, this is infeasible. While there are iterative alternatives to matrix decomposition, such as co-ordinate descent, conjugate gradient and dual co-ordinate descent, it may still be difficult to achieve high frame-rates with these.

We instead propose features of the special form $\psi[u] = e_{k[u]}$ where $e_i$ is a vector that is one at index $i$ and zero everywhere else, then the one-sparse inner product is simply a lookup $\beta^T\psi[u] = \beta^{k[u]}$ as in the PLT method described in the VOT13 challenge [26]. The particular type of features that we consider are quantised RGB colours, although a suitable alternative would be Local Binary Patterns. Recall from (7) that the histogram score can be considered an average vote. For efficiency, we therefore propose to apply linear regression to each feature pixel independently over object and background regions $\mathcal{O}$ and $\mathcal{B} \subset \mathbb{Z}^2$ using the per-image objective $\ell_{\text{hist}}(x, p, \beta) =$

$$\tfrac{1}{|\mathcal{O}|}\textstyle\sum_{u\in\mathcal{O}} \left(\beta^T\psi[u] - 1\right)^2 + \tfrac{1}{|\mathcal{B}|}\sum_{u\in\mathcal{B}} \left(\beta^T\psi[u]\right)^2 \tag{23}$$

where $\psi$ is short-hand for $\psi_{T(x,p)}$. Introducing the one-hot assumption, the objective decomposes into independent terms per feature dimension $\ell_{\text{hist}}(x, p, \beta) =$

$$\textstyle\sum_{j=1}^{M} \left[\frac{N^j(\mathcal{O})}{|\mathcal{O}|} \cdot (\beta^j - 1)^2 + \frac{N^j(\mathcal{B})}{|\mathcal{B}|} \cdot (\beta^j)^2\right] \tag{24}$$

where $N^j(\mathcal{A}) = |\{u \in \mathcal{A} : k[u] = j\}|$ is the number of pixels in the region $\mathcal{A}$ of $\phi_{T(x,p)}$ for which feature $j$ is non-zero $k[u] = j$. The solution of the associated ridge regression problem is

$$\beta^j_t = \frac{\rho^j(\mathcal{O})}{\rho^j(\mathcal{O}) + \rho^j(\mathcal{B}) + \lambda} \tag{25}$$

| | |
|---|---|
| Learning rate (template) $\eta_{\text{tmpl}}$ | 0.01 |
| Learning rate (histogram) $\eta_{\text{hist}}$ | 0.04 |
| Colour features | RGB |
| # bins colour histograms | $32 \times 32 \times 32$ |
| Merge factor $\alpha$ | 0.3 |
| Fixed area | $150^2$ |
| HOG cell size | $4 \times 4$ |

Table 1: The parameters we use for our experiments.

for each feature dimension $j = 1, \ldots, M$, where $\rho^j(\mathcal{A}) = N^j(\mathcal{A})/|\mathcal{A}|$ is the proportion of pixels in a region for which feature $j$ is non-zero. This expression has previously been used under probabilistic motivation [2, 33]. In the online version, the model parameters are updated

$$\begin{aligned} \rho_t(\mathcal{O}) &= (1-\eta_{\text{hist}})\rho_{t-1}(\mathcal{O}) + \eta_{\text{hist}}\rho'_t(\mathcal{O}) \\ \rho_t(\mathcal{B}) &= (1-\eta_{\text{hist}})\rho_{t-1}(\mathcal{B}) + \eta_{\text{hist}}\rho'_t(\mathcal{B}) \end{aligned} \quad (26)$$

where $\rho_t(\mathcal{A})$ is the vector of $\rho_t^j(\mathcal{A})$ for $j = 1, \ldots, M$.

### 3.5. Search strategy

When searching for the target's position in a new frame, we consider rectangular windows that vary in translation/scale but not aspect ratio/orientation. Rather than search jointly in translation/scale, we search first in translation and subsequently in scale. We follow Danelljan *et al.* [8] and learn a distinct, multi-scale template for scale search using a 1D Correlation Filter. This model's parameters are updated using the same scheme as the template learnt for translation. The histogram score is not suited to scale search because it will often prefer to shrink the target to find a window that is more purely foreground.

For both translation and scale, we search only in a region around the previous location. We also follow prior works that adopt Correlation Filters for tracking [18, 8] in using a Hann window during search as well as for training. Together these can be considered an implicit motion model.

The size of the translation template is normalised to have a fixed area. This parameter can be tuned to trade tracking quality for speed, as shown in the following section.

## 4. Evaluation

We compare Staple to competing methods on two recent and popular benchmarks, VOT14 [24, 25] and OTB [39], and demonstrate state-of-the-art performance. To achieve an up-to-date comparison, we report the results of several recent trackers in addition to the baselines that are part of each benchmark, using the authors' own results to ensure a fair comparison. Therefore, for each evaluation, we can only compare against those methods that provide results for it. To aid in reproducing our experiments, we make the source code of our tracker and our results available on our website: www.robots.ox.ac.uk/~luca/staple.html.

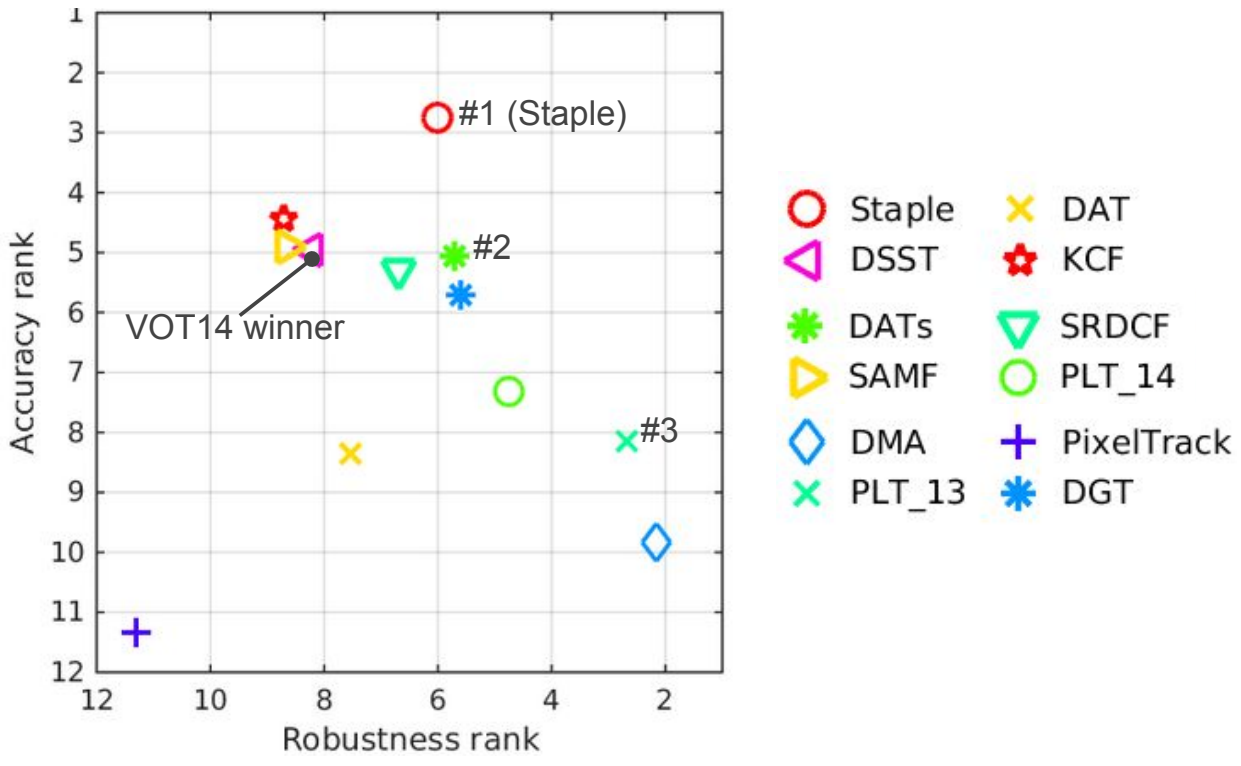

Figure 3: Accuracy-Robustness rank plot for `report_challenge`. Better trackers are closer to the top right corner.

In Table 1, we report the values of the most important parameters we use. Contrary to standard practice, we do not choose the parameters of the tracker from the testing set, but instead use VOT15 as a validation set.

### 4.1. VOT14 and VOT15

**The benchmark.** VOT14 [24] compares competing trackers on 25 sequences chosen from a pool of 394 to represent several challenging situations: camera motion, occlusion, illumination, size and motion change. Two performance measures are used. The *accuracy* of a tracker on a sequence is expressed as the average per-frame overlap between its predicted bounding box $r_t$ and the ground truth $r_{\text{GT}}$ using the intersection-over-union criterion $S_t = \frac{|r_t \cap r_{\text{GT}}|}{|r_t \cup r_{\text{GT}}|}$. The *robustness* of a tracker is its number of failures over the sequence, with a failure determined to have occurred when $S_t$ becomes zero. Since the focus of the benchmark is on *short-term* tracking, a tracker that fails is automatically reinitialised to the ground truth five frames after the failure.

Given the nature of the two performance measures, it is crucial to consider them jointly. Considering either in isolation is uninformative, since a tracker that fails frequently will be re-initialised more often and likely achieve higher accuracy, while zero failures can always be achieved by reporting that the object occupies the entire video frame.

**Results.** To produce Table 2 and Figure 3, we used the most recent version of the VOT toolkit available at submission time (commit `d3b2b1d`). From VOT14 [25] we only include the top performers: DSST [8], SAMF [29], KCF [18], DGT [6], PLT_14 and PLT_13. Table 2 reports the average accuracy and number of failures for each tracker, together with an overall ranking devised from both. Figure 3 visualises independent ranks for each metric on two axes. Surprisingly, our simple method significantly outperforms all VOT14 entries, together with many recent trackers published after VOT14. In particular, it surpasses trackers like Correlation Filter-based DSST [8], SAMF [29] and KCF [18], colour-based PixelTrack [11], DAT, DATs (with scale) [33] and DGT [6], and also more complex and far slower methods like DMA [40] and SRDCF [9], which operates below 10 FPS. It is interesting to observe how Staple performs in comparison to the second-best correlation and colour trackers, SRDCF and DATs: it achieves a 7% improvement in accuracy and 41% improvement in number of failures over SRDCF, and an 11% improvement in accuracy and 13% improvement in number of failures over DATs. Considering the two metrics individually, Staple is by far the best method in terms of accuracy and the fourth for number of failures, after DMA, PLT_13 and PLT_14. However, all these trackers perform poorly in terms of accuracy, scoring at least 20% worse than Staple.

| Tracker | Year | Where | Accuracy | # Failures | Overall Rank |
|---|---|---|---|---|---|
| Staple (proposed) | - | - | 0.644 | 9.38 | 4.37 |
| DATs [33] | 2015 | CVPR | 0.580 | 13.17 | 5.39 |
| PLT_13 [24] | 2013 | VOT | 0.523 | 1.66 | 5.41 |
| DGT [6] | 2014 | TIP | 0.534 | 13.78 | 5.66 |
| SRDCF [9] | 2015 | ICCV | 0.600 | 15.90 | 5.99 |
| DMA [40] | 2015 | CVPR | 0.476 | 0.72 | 6.00 |
| PLT_14 [24] | 2014 | VOT | 0.537 | 3.41 | 6.03 |
| KCF [18] | 2015 | PAMI | 0.613 | 19.79 | 6.58 |
| DSST [8] | 2014 | BMVC | 0.607 | 16.90 | 6.59 |
| SAMF [29] | 2014 | ECCVw | 0.603 | 19.23 | 6.79 |
| DAT [33] | 2015 | CVPR | 0.519 | 15.87 | 7.95 |
| PixelTrack [11] | 2013 | ICCV | 0.420 | 22.58 | 11.31 |

Table 2: Ranked results for VOT14. First, second and third entries for accuracy, number of failures (over 25 sequences) and overall rank are reported. Lower rank is better.

| Tracker | Accuracy | # Failures |
|---|---|---|
| Staple (proposed) | 0.538 | 80 |
| DSST [8] | 0.491 | 152 |
| DATs [33] | 0.442 | 124 |

Table 3: Ranked result for the 60 sequences of VOT15.

For completeness, we also present results for VOT15 in Table 3, comparing Staple against the second-best performer in Table 2 (DATs) and the winner of VOT14 (DSST). Our performance is significantly better than DATs and DSST in terms of both accuracy (respectively +22% and +10%) and robustness to failures (+35% and +47%).

In this experiment, we have kept the hyper-parameters that were chosen for VOT15. However, this is in accord with convention, since the VOT benchmark has never included a validation set on the assumption that the hyper-parameters would be simple enough not to vary significantly between datasets.

### 4.2. OTB-13

**The benchmark.** As with VOT, the idea of OTB-13 [39] is to evaluate trackers on both accuracy and robustness to failure. Again, prediction accuracy is measured as intersection over union between the tracker's bounding box and ground truth. A success is detected when this value is above a threshold $t_o$. In order not to set a specific value for such a threshold, the area under the curve of success rates at different values of $t_o$ is used as a final score.

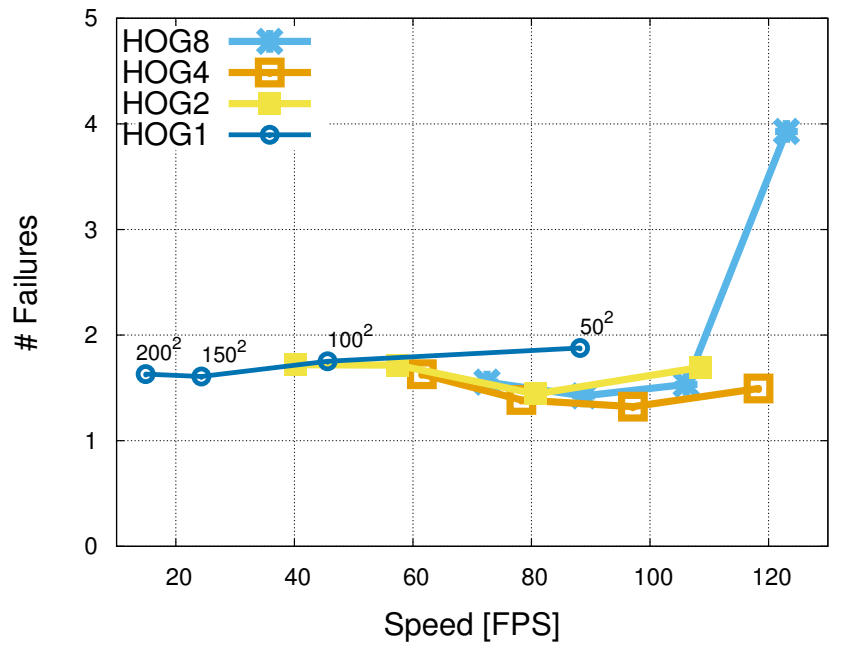

Figure 5: Number of failures (lower is better) in relation to speed for HOG cells of size $1 \times 1$, $2 \times 2$, $4 \times 4$ and $8 \times 8$.

**Results.** Our results for OTB have been obtained using exactly the same code and parameters used for VOT14/15. The only difference is that we are constrained to use one-dimensional histograms for the few grayscale sequences present in the benchmark. Figure 4 reports the results of OPE (one pass evaluation), SRE (spatial robustness evaluation) and TRE (temporal robustness evaluation). Staple performs significantly better than all the methods reported in [39], with an average relative improvement of 23% with respect to the best tracker evaluated in the original benchmark (Struck [16]). Moreover, our method also outperforms recent trackers published after the benchmark such as MEEM [41], DSST [8], TGPR [13], EBT [38] and also trackers that make use of deep conv-nets like CNN-SVM [19] and SO-DLT [37], while running at a significantly higher frame rate. The only comparable method in terms of frame-rate is ACT [10], which however performs substantially worse in all the evaluations. Since ACT learns a colour template using Correlation Filters, this result shows that the improvement that Staple achieves by combining template and histogram scores cannot be attributed solely to the introduction of colour. On OTB, the only tracker performing better than Staple is the very recent SRDCF [9]. However, it performs significantly worse on VOT14. Furthermore, it has a reported speed of only 5 FPS, which severely limits its applicability.

### 4.3. Efficiency

With the above reported configuration, our MATLAB prototype runs at approximately 80 frames per second on a machine equipped with an Intel Core i7-4790K @4.0GHz. However, it is possible to achieve higher frame rates with a relatively small loss in terms of performance, by adjusting the size of the patch from which the models are computed. For example (refer to Figure 5), using HOG cells of size $2 \times 2$ and a fixed area of $50^2$ causes only a small increase in the number of failures, yet boosts the speed beyond 100 frames per second. The accuracy follows a similar trend.

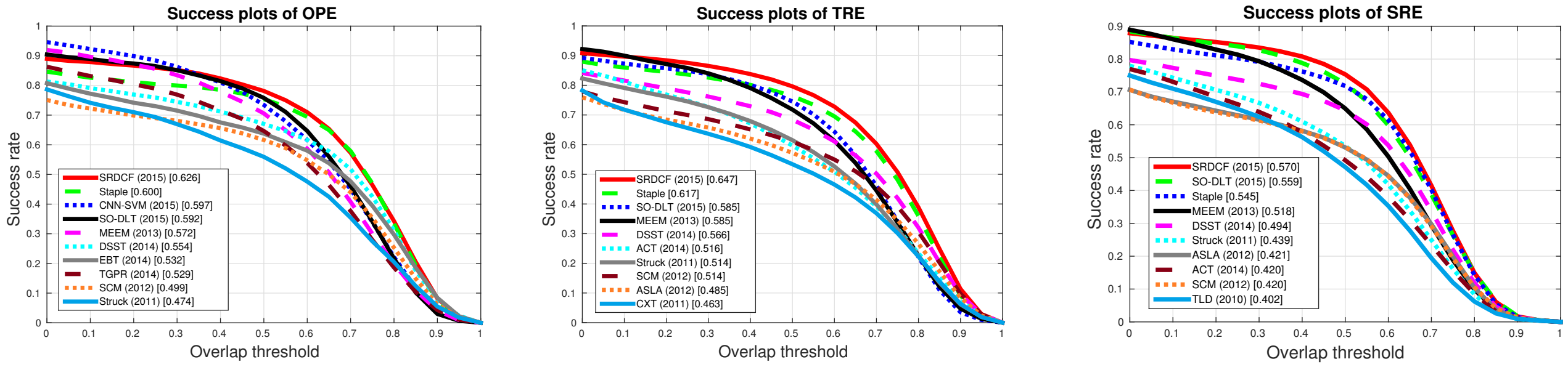

Figure 4: Success plots for OPE (one pass evaluation), TRE (temporal robustness evaluation) and SRE (spatial robustness evaluation) on the OTB-13 [39] benchmark.

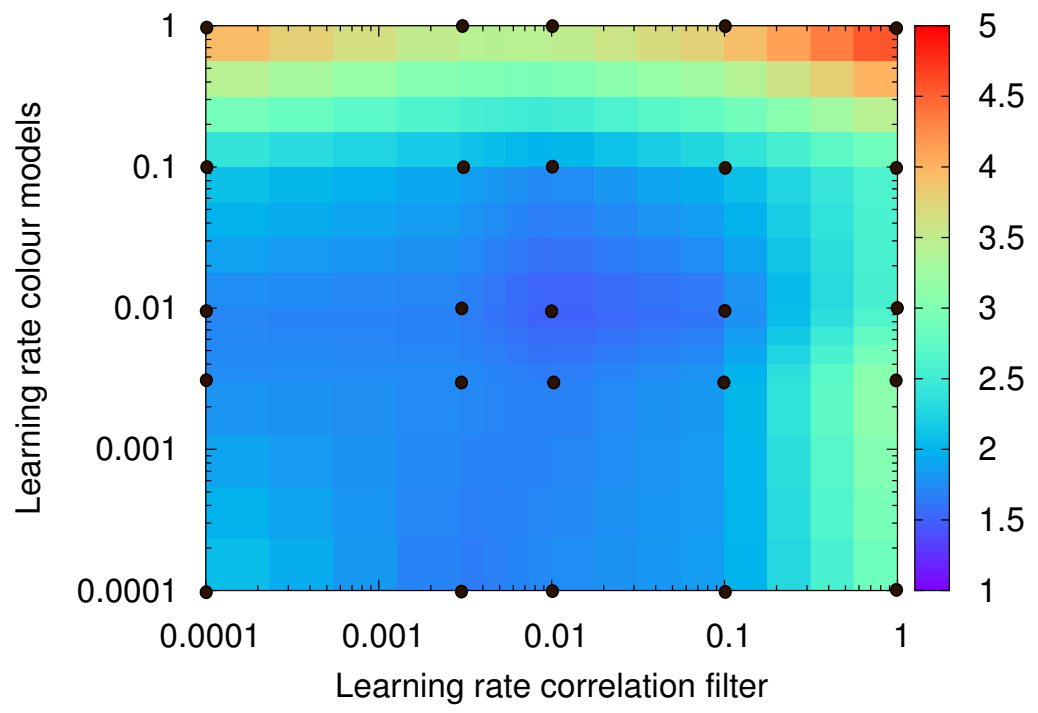

Figure 6: Number of failures (lower is better) in relation to the learning rates $\eta_{\text{tmpl}}$ and $\eta_{\text{hist}}$. Black points were obtained experimentally, others were interpolated.

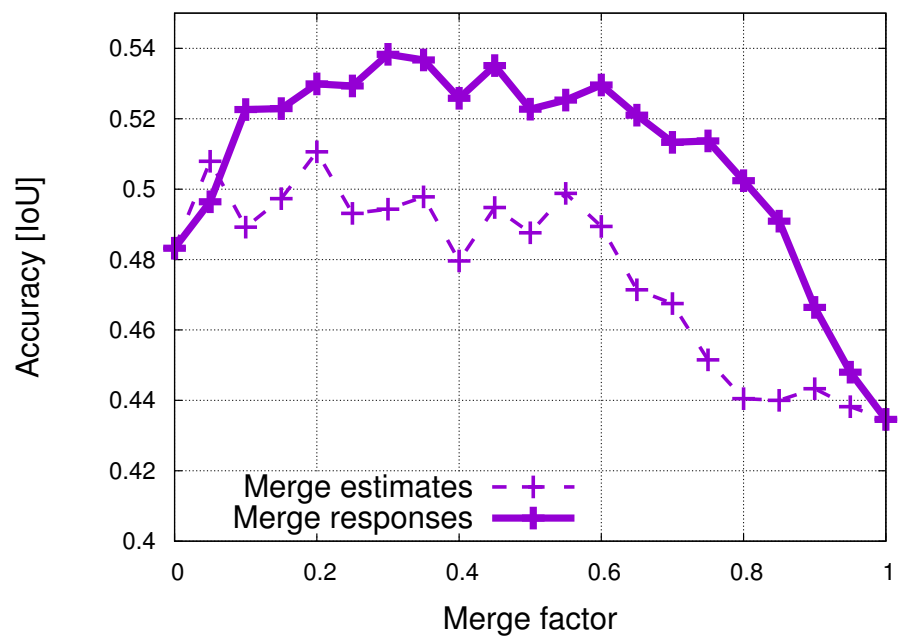

Figure 7: Accuracy (higher is better) vs. merge factor $\alpha$.

### 4.4. Learning rate experiments

The learning rates $\eta_{\text{tmpl}}$ and $\eta_{\text{hist}}$, used respectively for the template (21) and histogram (26) model updates, determine the rate at which to replace old evidence from earlier frames with new evidence from the current frame. The lower the learning rate, the higher the relevance given to model instances learnt in earlier frames. The heatmap of Figure 6 illustrates how maximal robustness is achieved at around 0.01 for both $\eta_{\text{tmpl}}$ and $\eta_{\text{hist}}$. The accuracy follows a similar trend.

### 4.5. Merge factor experiments

In Figure 7, we show how the accuracy of Staple is significantly influenced by the choice of the merge factor $\alpha$ that regulates $\gamma_{\text{tmpl}}$ and $\gamma_{\text{hist}}$ in (3): the best performance is achieved around $\alpha = 0.3$. The robustness follows a similar trend. Figure 7 also shows that the strategy of merging the dense responses of the two ridge regression problems achieves significantly better performance than a mere interpolation of the final estimates would, suggesting that choosing models with compatible (and complementary) dense responses is a winning choice.

## 5. Conclusion

By learning their model from circular shifts of positive examples, correlation filters fail to learn a component that is invariant to permutations. This makes them inherently sensitive to shape deformation. We therefore propose a simple combination of template and histogram scores that are learnt independently to preserve real-time operation. The resulting tracker, Staple, outperforms significantly more complex state-of-the-arts trackers in several benchmarks. Given its speed and simplicity, our tracker is a logical choice for applications that require computational effort themselves, and in which robustness to colour, illumination and shape changes is paramount.

**Acknowledgements.** This research was supported by Apical Imaging, Technicolor, EPSRC, the Leverhulme Trust and the ERC grant ERC-2012-AdG 321162-HELIOS.